\documentclass[conference]{IEEEtran}
\IEEEoverridecommandlockouts
\usepackage{comment}
\usepackage{amsmath,amssymb,amsfonts}
\usepackage{algorithmic}
\usepackage{graphicx}
\usepackage{textcomp}
\usepackage{xcolor}
\usepackage{graphicx}
\usepackage{amsmath}
\usepackage{cite}
\usepackage{wrapfig}
\usepackage{soul}

\usepackage[colorlinks,urlcolor=blue,linkcolor=blue,citecolor=blue]{hyperref}

\usepackage{mdframed}
\usepackage{tcolorbox}
\def\BibTeX{{\rm B\kern-.05em{\sc i\kern-.025em b}\kern-.08em
    T\kern-.1667em\lower.7ex\hbox{E}\kern-.125emX}}
\begin{document}

\title{Neurosymbolic Value-Inspired AI (Why, What, and How)}

\author{\and
\IEEEauthorblockN{\textbf{Amit Sheth}}
\IEEEauthorblockA{Artificial Intelligence Institute\\
University of South Carolina\\
Columbia, SC, USA \\
\texttt{amit@sc.edu}}
\and
\IEEEauthorblockN{\textbf{Kaushik Roy}}
\IEEEauthorblockA{Artificial Intelligence Institute\\
University of South Carolina\\
Columbia, SC, USA \\
\texttt{kaushikr@email.sc.edu}}
}

\maketitle
\begin{abstract}
The rapid progression of Artificial Intelligence (AI) systems, facilitated by the advent of Large Language Models (LLMs), has resulted in their widespread application to provide human assistance across diverse industries. This trend has sparked significant discourse centered around the ever-increasing need for LLM-based AI systems to function among humans as part of human society, sharing human values, especially as these systems are deployed in high-stakes settings (e.g., healthcare, autonomous driving, etc.). Towards this end, neurosymbolic AI systems are attractive due to their potential to enable easy-to-understand and interpretable interfaces for facilitating value-based decision-making, by leveraging explicit representations of shared values. In this paper, we introduce substantial extensions to Khaneman's System one/two framework and propose a neurosymbolic computational framework called Value-Inspired AI (VAI). It outlines the crucial components essential for the robust and practical implementation of VAI systems, aiming to represent and integrate various dimensions of human values. Finally, we further offer insights into the current progress made in this direction and outline potential future directions for the field.
\end{abstract}
\section{Why does AI need a Value System?}
Since the inception of AI systems, a primary goal has been their seamless integration into human society, aiming to assist in demanding tasks, such as large-scale automation. Consequently, discussions about their responsible utilization, particularly as they gain advanced capabilities, have been integral to active and interdisciplinary academic discourse. Questions have arisen about the values embedded in these systems and, more broadly, how to ensure their usage benefits humankind. In recent times, the exceptional capabilities of LLMs in AI have accelerated the widespread adoption of AI across diverse industries. However, this adoption has not been without profound social consequences for human users, giving rise to unforeseen social risks like biases and ethical concerns. In response to these risks, there is an urgent call to implement ``controls'' on LLMs and their outcomes \cite{weidinger2021ethical}. For humans functioning within a society, the basis for providing such ``controls'' on their functioning is rooted in a set of shared values that enjoy broad consensus among the populace. These values span various dimensions, encompassing ethics, socio-cultural norms, policies, regulations, laws, and other pertinent aspects \cite{schwartz1987toward}. Due to the complexity of such a value-based framework, decision-making behaviors in these situations are usually established by trained personnel in government positions with support from legislation on behalf of the wider population within society \cite{leal2016role}. We make an important distinction between these carefully thought-out, expert-defined societal values for the synergistic functioning of humans within society, and preference patterns of a large collective of people, which may itself contain implicit notions of values, but may also consequently contain population-wide biases. The latter does not readily clarify an unambiguous value-based stance that considers maintaining societal order and is, therefore, not suited for incorporation within VAI systems without additional audits and checks by regulatory bodies. 


\begin{tcolorbox}[title=Why VAI?,float,floatplacement=htb]
\textit{As AI systems increasingly take center stage in human assistance, we should expect the system to be aware of the values that a human operator would be aware of and adhere to. For example, an AI for driving assistance must be aware of the values of a human driver (e.g., values pertaining to driving rules, regulations, policies, and ethical behaviors laid out by the appropriate branches of governance), and abide by those values. }
\end{tcolorbox}

\subsection{AI-assited Driving - Motivating VAI Systems}\label{sec:examples}
The following scenario represents a case where a clearly defined value system is necessary for decision-making and ensuring subsequent compliance with respect to acceptable values within human societies (e.g., socio-cultural norms, policies, regulations, laws, etc. \cite{purohit2020knowledge}). The example is extremely challenging and intended to motivate the imperative need for VAI frameworks. Consider a variant of the classic thought experiment known as the “trolley problem” in the AI-assisted driving setting that asks: Should you pull aside to divert your runaway vehicle so that it kills one person rather than five? Alternatively, what if a bicycle suddenly enters the lane? Should the vehicle swerve into oncoming traffic or hit the bicycle? What choices are moral in these scenarios? One approach can be to decide based on values prioritizing society, i.e., the fewest deaths, or a solution that values individual rights (such as the right not to be intentionally put in harm’s way) \cite{geisslinger2021autonomous}. Ultimately, decisions are subject to a clearly defined and specified value system so that AI systems can be objectively and responsibly managed.

\section{What is the Role of Neurosymbolic AI for VAI?}\label{sec:what}
\begin{figure*}[!htb]
    \centering
    \includegraphics[width=\linewidth,trim = 0cm 2cm 0cm 0cm, clip]{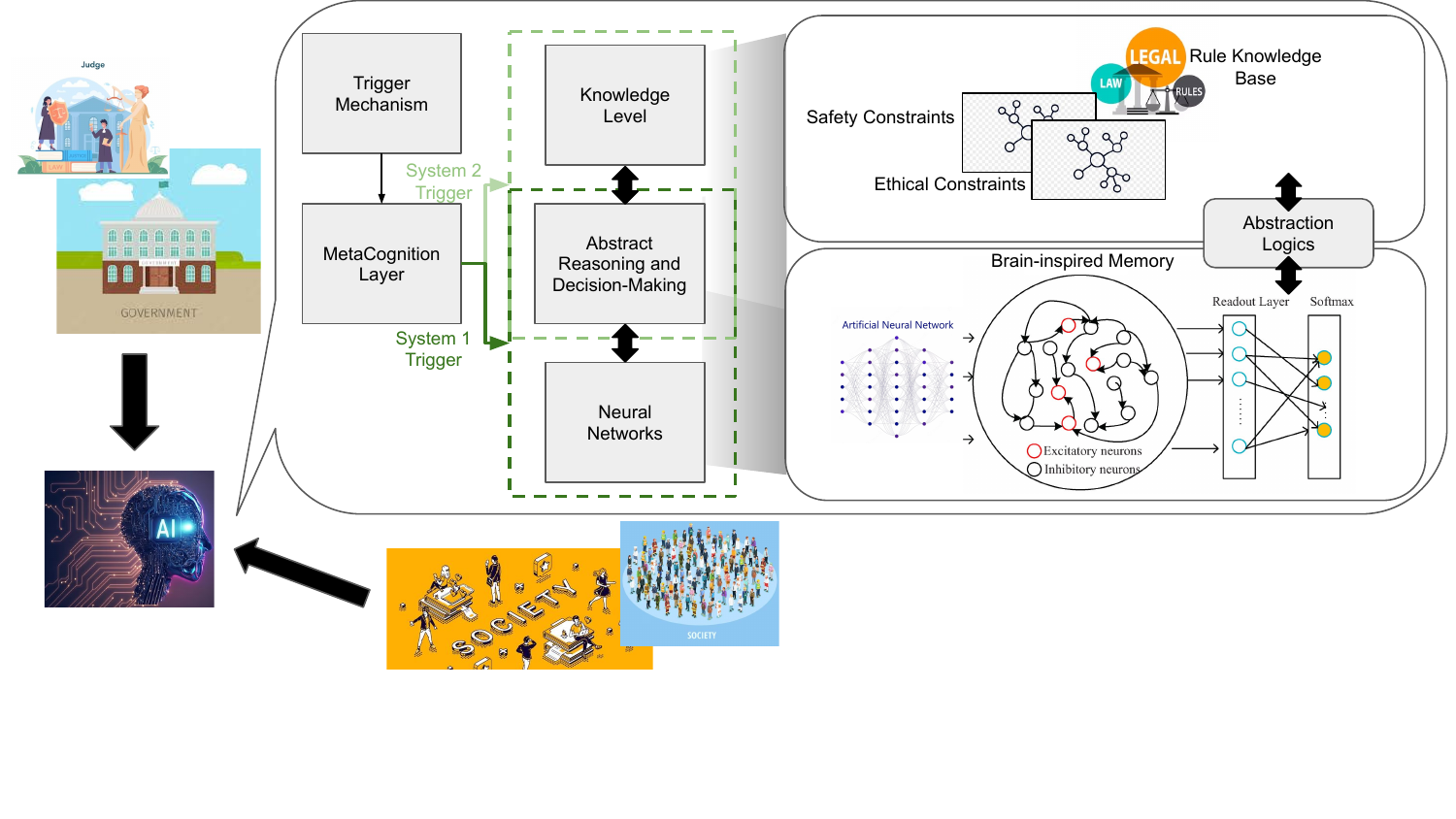}
    \caption{An Illustration of the components of our proposed VAI framework. Components include a knowledge level (see Section \ref{subsec:kg}, neural networks (see Section \ref{subsec:dynNN}, abstraction logics, reasoning, and decision making (see Section \ref{subsec:ab_l}), and metacognition layers with triggers (see Section \ref{subsec:meta}. We explain that a robust VAI method will need to integrate all these components using a neurosymbolic AI framework (see Section \ref{subsec:nesy}).}
    \label{fig:arch}
\end{figure*}
\paragraph*{The Adequacy of Daniel Khaneman's Framework for VAI} As exemplified in Section \ref{sec:examples}, the mechanisms for value-inspired decision-making in real-world scenarios can get quite complex and nuanced. Figure \ref{fig:arch} shows a general architecture consisting of the essential components for human value-based decision-making. We borrow from and extend Khaneman's system one and two frameworks, which has been established as a gold standard for formulating AI system objectives toward achieving human-like decision-making outcomes \cite{kahneman2011thinking}. Figure \ref{fig:arch} illustrates the two systems. System two functions involve the representation of deliberative ``thought structures'' (e.g., values, i.e., social-cultural-ethical norms, represented as graphs in the figure, and laws, rules, policies, and regulations, represented as a rule knowledge base in the figure), and reasoning over such structures. System one involves faster and reflexive elements guided by familiar patterns, depicted using neural networks in the figure. Neural networks are statistical methods that primarily rely on pattern recognition-based decision-making. Traditionally, System two components lend themselves better to \textit{symbolic} representations, and System one components lend themselves better to sub-symbolic or \textit{neural} representations (in the contemporary sense). Thus, it is natural to combine the two using a unified \textit{neurosymbolic} framework \cite{hitzler2022neuro,garcez2023neurosymbolic}. However, as seen in the remainder of Figure \ref{fig:arch}, Khaneman's framework still needs extension due to its lack of specificity regarding key requirements for a complete VAI architecture. We elucidate these specifics in the following subsections.
\subsection{Robust and Dynamic Knowledge Representation of Values}\label{subsec:kg}
Current realizations of the System two part of Khaneman's framework, lacks definitions of three primary facets of a value system required for the synergistic functioning of humans under a shared interpretation of values, especially in a society as diverse as humankind - (f1) \textit{Unambiguous}, (f2) \textit{Dynamic}, (f3) \textit{Rationalizable}. \textit{Unambiguous} refers to a well-defined and precise interpretation of shared values, typically achieved through iterative procedures based on consensus among interested parties. \textit{Dynamic} refers to the malleable nature of human values. Established human values do not change easily. However, they are subject to change depending on changing societal contexts, and in most cases, they do so organically without much hindrance to structures that are already in place. \textit{Rationalizable} - the core axioms for the definitions in (1) are mechanisms of change in (2) are grounded in principles of rational decision-making, verifiable through some form of auditing or scrutiny (e.g., everybody can read the constitution and law books). We propose knowledge graphs as a viable symbolic representation of values that meet the requirements for (f1), (f2), and (f3).

The semantic web community has extensively dealt with diverse and heterogeneous data sources and successfully integrated them into high-utility knowledge representations, such as ontologies and knowledge graphs (e.g., Google Knowledge Graph, Wikipedia, etc.) \cite{sheth2019knowledge}. This ecosystem, centered around knowledge graphs, has proven its robustness in meeting information needs, specifically in regards to (f1), (f2), and (f3), across dimensions of quality, scale, and dynamic content changes. Facet (f1) is achieved through processes for ensuring \textit{Ontological Commitment}. Facet (f2) is often a consequence of ensuring (f1), by which mechanisms to changes in the ontology design patterns and their practical implementation, both at the instance-level and ontology-level, are considered. Facet (f3) is ensured through an established body of theoretical results on the aspects of rationality in machines, namely -soundness, completeness, verifiability, and decidability of knowledge graphs and knowledge graph-based reasoners \cite{ter2005combining}. Today, knowledge graphs play a central role in various information processing and management tasks, including semantically-enriched applications such as the web and other AI systems for search, browsing, recommendation, advertisement, and summarization in diverse domains. 
\subsection{Brain-Inspired Memory Structures}\label{subsec:dynNN}
Transformer-based neural network architectures in LLMs have recently come to dominate the space of implementations for System one in Khaneman's framework. Established work on Brain-inspired cognitive architectures have made clear distinctions between different types of memories, based on their perceived roles. For example, declarative memory captures unchanging facts about the world, and episodic memory retains information about deviants from common schemas (e.g., birds that can't fly) \cite{langley2009cognitive}. This plays a big part in several humans co-existing based on a common set or shared interpretation of values. One or more individuals may not like specific aspects of another's values, but this can be safely ignored as part of an episodic deviant as long as it is not a significant deal breaker.   

We first observe that systems such as LLMs are fundamentally large neural networks. Therefore, their loss surfaces, for most traditional choices of the loss function, are highly non-convex, resulting in complex parameter-space dynamics, or \textit{parametric-memory} dynamics. However, current training methods do not lead to models that adequately capture the dynamics. Specifically, the models do not possess \textit{dynamic working memories}, and instead solve for a fixed point, i.e., a \textit{static working memory}, that is invoked at inference time. The resulting inferences are expected to \textit{generalize} to unseen test cases. We argue that in order to plausibly interact with the dynamic nature of the symbolic value representations, neural network architectures will need re-designs to capture a reasonable notion of \textit{state dynamics}, thus making a distinction between episodic memories (a sequence of states), and generalizable memory (individual-episode-agnostic generalizable patterns). Furthermore, since the neural network will need to interact with the symbolic layer, these changes will allow the network structures to have corresponding interfaces with appropriate parts of the symbolic layer. For example, generalizable memory interacts with generalized ontological constructs, and episodic memory interacts with instance-level changes or lack of conformance to the higher-level ontological constructs. 
\subsection{Temporal Abstraction Logics}\label{subsec:ab_l}
Although Khaneman's framework defines individual types of systems based on their functions, it does not define how these systems are to communicate with clarity. What might be a suitable modality, logic, or knowledge representation to enable such communication? To draw this bridge, we will individually examine neural network-based knowledge representations and more classical symbolic knowledge representations and propose a solution. Neural networks are adept at pattern recognition and consequently excel at capturing linguistic structure in the training data. 
Furthermore, The recent success of LLMs, particularly in benchmark tasks that test common sense, such as the Winograd challenge, shows evidence that \textit{semantic} understanding does not always require a strictly structured propositional description. Rather, that semantic knowledge is deeply coupled with language patterns (e.g., linguistic and syntactic). At the same time, LLMs have performed embarrassingly poorly at tasks that require demonstration of several other aspects of common sense understanding (e.g., intuitive physics, planning and causal sequence capture), thus showing a lack of a ``full-bodied'' understanding of the world. The holy grail of a full-bodied understanding of the world has been to adequately capture the full breadth of \textit{relationships} between linguistic comprehension and semantic knowledge, one of the early endeavors of symbolic AI, e.g., WordNet (representing linguistic variations and word senses), ConceptNet (relationships between linguistic variations, word senses, and broader concepts and their properties), WikiData (relationships between concepts and entities such as people, places, things, and organizations), etc \cite{browning2023language}. 
\begin{figure*}[!htb]
    \centering
    \includegraphics[trim = 0.2cm 3.4cm 0cm 0cm, clip, width=\linewidth]{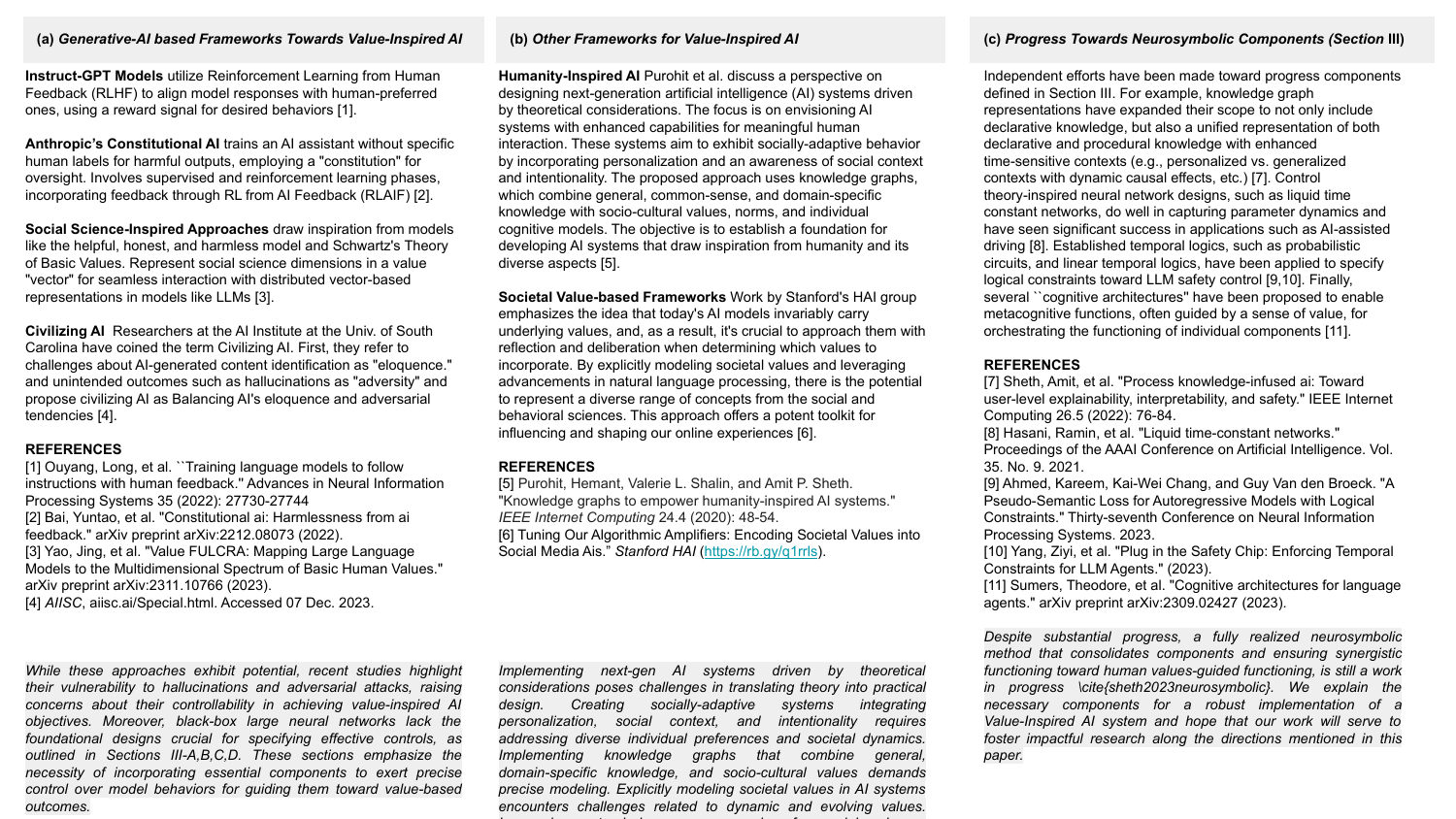}
    \caption{Summary of our Progress toward VAI. (a) Talks about Generative AI-based conceptualizations of VAI frameworks by other researchers. (b) Talks about other frameworks that explicitly talk about integrating human values in society with AI systems. (c) Talks about progress towards the implementation of individual components mentioned in Sections \ref{subsec:kg}, \ref{subsec:dynNN}, \ref{subsec:ab_l}, and \ref{subsec:meta}.}
    \label{fig:how}
\end{figure*}
Therefore, we believe that the answer to a neurosymbolic interface for communication between Systems one and two lies in a new kind of knowledge representation which we will refer to as \textit{abstraction logics}, that is a mix of the classical propositional expressions and more distributional representations. Note that we are proposing a \textit{blended} representation - this differs from previous work on neurosymbolic methods that either extract propositional representations from neural mechanisms or compress propositional information into continuous-valued vector spaces for consumption by neural networks. Moreover, existing neurosymbolic methods have primarily concentrated on a static portrayal of the world. In contrast, the abstraction logics developed in our framework must adeptly handle the dynamic nature of knowledge representations in System two and the dynamical model of neural networks proposed for System one. This necessitates incorporating temporal aspects to capture the evolving nature of information. For instance, many value systems incorporate temporal dimensions (e.g., carbon emission goals for countries and corporations).
We argue that \textit{relationships} expressed using these new kinds of \textit{abstraction logics}, will be at the heart of transitioning from machine comprehension of a string of sounds or letters to an internal representation of ``meaning'', that serves to enhance clarity in both AI and human interpretations of shared values.

\subsection{Metacognition Layers and Triggers}\label{subsec:meta}

Finally, Khaneman's framework does not specify when to invoke either system. Human decision-making and activities related to Systems one and two are context-specific and result in either reflexive decisions and actions based on prior experiences, or deliberative processes that involve slower thinking and reasoning. In the AI-assisted driving scenario discussed in Section \ref{sec:examples}, most decisions occur reflexively, guided by repetitive driving patterns. More intricate decisions, however, necessitate a deliberative process that considers the value structures for careful thought regarding subsequent steps and actions. Such decisions regarding when to choose between intricate vs. reflexive action form aspects of \textit{metacognition}, where a computational mechanism \textit{triggers} either System one or two, or both, depending on the specific task \cite{cox2005metacognition}. This computational mechanism must be rooted in the fundamental interface between Systems one and two, namely, the abstraction logics introduced in Section \ref{subsec:ab_l}. Therefore, a practical implementation of a VAI architecture requires a module with functions geared toward implementing criteria that determine when Systems one and two are invoked. This module is crucial in orchestrating the interplay between reflexive and deliberative decision-making processes, facilitating the necessary dynamic and context-sensitive AI system responses.

\subsection{Neurosymbolic VAI: Putting it All Together}\label{subsec:nesy}
As covered in Sections \ref{subsec:kg}, \ref{subsec:dynNN}, \ref{subsec:ab_l}, and \ref{subsec:meta}, the synthesis of a comprehensive and capable neurosymbolic, VAI architecture can be achieved by integrating the components mentioned above. Each component is equipped with specific implementations tailored to its unique functions: a dynamic knowledge graph-centered network and reasoning mechanisms for robustly representing values and facilitating decision-making based on these values (System two); brain-inspired neural network-based dynamical systems for the expressive encoding of various aspects of memory (System one); temporal abstraction logics to facilitate the interaction between Systems one and two; and metacognition layers and triggers orchestrating the overall functioning of the AI system. This integrated architecture enables a computational framework capable of harmonizing diverse cognitive processes and enhancing the AI system's adaptability and responsiveness during operation with humans in human society, under a shared value system.

\section{How Far Along We Are and Future Directions}
Figure \ref{fig:how} provides an overview of the progress that researchers have made so far in incorporating human values within AI systems. 
\subsection{Generative-AI-based Frameworks Towards VAI}
Figure \ref{fig:how} (a) describes techniques used by three main classes of techniques for the incorporation of human values, specifically incorporation by aligning model outputs with human preferences. While these approaches exhibit potential, recent studies highlight their vulnerability to hallucinations and adversarial attacks, raising concerns about their controllability in achieving VAI. Moreover, black-box large neural networks lack the foundational designs crucial for specifying effective controls, as outlined in Sections \ref{subsec:kg}, \ref{subsec:dynNN}, \ref{subsec:ab_l}, and \ref{subsec:meta}. These sections emphasize the necessity of incorporating essential components to exert precise control over model behaviors for guiding them toward value-based outcomes.
\subsection{Other Frameworks for VAI}
Figure \ref{fig:how} (b) shows previous efforts that have characterized the complexity of human values within society. These works emphasize key challenges related to the dynamic and evolving nature of explicitly modeling societal values in AI systems and propose knowledge graphs and LLMs as candidates to handle such dynamics. Adapting LLM, or general neural network processing techniques to incorporate values represented in KGs, requires a computing framework for forming a clear understanding of value-based perspectives and considerations within the neural network's internal structures. These works have not concretely talked about such a framework. In this work, we provide a road map with concrete steps and specific implementation strategies for achieving VAI through a neurosymbolic computational method.
\subsection{Progress Towards Neurosymbolic Components (Section \ref{sec:what})}
Figure \ref{fig:how} (c) shows substantial progress across all components defined in Section \ref{sec:what}. Despite advances in the individual components, a fully realized neurosymbolic method that consolidates components and ensures synergistic functioning toward human values-guided functioning is still a work in progress \cite{sheth2023neurosymbolic}. 

\section*{Concluding Remarks}
In this paper, we introduce VAI systems and expand on Khaneman's System one/two framework by providing detailed outlines of components necessary for the robust implementation of VAI systems. Specifically, we identify existing implementation challenges and present a clear road map for integrating explicit models of societal values in knowledge graphs, paired with technical advances in neural methods, abstraction logics, and metacognition methods, within a neurosymbolic computational framework. We hope our work will inspire building on the considerable progress and further stimulate impactful research along the directions mentioned in this paper.

\section*{Acknowledgements}
This research was supported in part by NSF Awards 2335967 "EAGER: Knowledge-guided neurosymbolic AI" with guardrails for safe virtual health assistants" (opinions are those of authors and not the sponsor), and feedback from AIISC colleagues.
\section*{Authors}

\noindent \textbf{Amit Sheth} is the founding director of the AI Institute of South Carolina (AIISC), NCR Chair, and a professor of Computer Science and Engineering at the University of South Carolina. He received the 2023 IEEE-CS Wallace McDowell award and is a fellow of IEEE, AAAI, AAIA, AAAS, and ACM, and has published several articles and papers at reputed venues in the area of this article \cite{sheth2021knowledge,sheth2022process,sheth2023neurosymbolic}. Contact him at: \texttt{amit@sc.edu} 

\noindent \textbf{Kaushik Roy}
is a Ph.D. student at the AIISC. His research interests include the development of algorithms to combine statistical data-driven learning mechanisms with curated domain-specific information from external knowledge sources. He has published his work at reputed venues and has an active publication record in the area of this article \cite{roy2021knowledge,roy2022tutorial,roy2022tdlr,roy2023proknow} . Contact him at: \texttt{kaushikr@email.sc.edu} 
\bibliographystyle{IEEEtran}
\bibliography{references.bib}
\end{document}